\documentclass[runningheads]{llncs}

\usepackage[mobile]{eccv}

\usepackage{eccvabbrv}
\usepackage{graphicx}
\usepackage{booktabs}
\usepackage[accsupp]{axessibility}

\usepackage{siunitx} % Added by Ted
\usepackage{pifont} % Added by Ted
\usepackage{stfloats} % Added by Ted
\usepackage{multirow} % Added by Ted
\usepackage[table]{xcolor} % Added by Santi

\definecolor{citationblue}{RGB}{0, 113, 188} % Added by Ted
\newcommand{\cmark}{\ding{51}} % Added by Ted
\newcommand{\xmark}{\ding{55}} % Added by Ted
\newcommand{\rurl}[1]{\href{http://#1}{\nolinkurl{#1}}} % Added by Ted
\newcommand{\finding}[1]{\par\vspace{2pt}\noindent\fcolorbox{gray!45}{gray!8} % Added by Ted
{\parbox{\dimexpr\linewidth-2\fboxsep-2\fboxrule}{\textbf{#1}}}\par\vspace{3pt}\noindent} % Added by Ted

\makeatletter
\def\ps@firstpage{%
  \let\@oddhead\@empty
  \let\@evenhead\@empty
  \setlength{\footskip}{18pt}%
  \def\@oddfoot{\hfil\small\textit{European Conference on Computer Vision (ECCV 2026) - DriveX}\hfil}%
  \let\@evenfoot\@oddfoot
}
\makeatother

\usepackage[pagebackref,breaklinks,colorlinks,allcolors=eccvblue]{hyperref}

\begin{document}

\title{Emergent 3D Instance Segmentation from Self-Supervised Point Transformers}
\titlerunning{TokenGraph3D}

\author{Ted Lentsch\inst{1} \and Santiago Montiel-Mar{\'\i}n\inst{2} \and Holger Caesar\inst{1} \and Julian F.P. Kooij\inst{1}}
\authorrunning{T.~Lentsch et al.}

\institute{
    Department of Cognitive Robotics, Delft University of Technology, The Netherlands\\
    \and
    Department of Electronics, University of Alcal{\'a}, Spain
}

\maketitle
\thispagestyle{firstpage}

\begin{abstract}
    Unsupervised 3D instance segmentation of outdoor LiDAR scans has traditionally relied on handcrafted geometric priors such as density-based clustering, motion cues, or projected 2D detections. In this work, we investigate whether a frozen, self-supervised point transformer already contains the structural information required to isolate object instances without any handcrafted geometric prior. Using this transformer purely as a feature extractor, we probe its internal representations across the SemanticKITTI, nuScenes, and Waymo Perception datasets. Our analysis yields four core insights: (1) the instance signal concentrates in the attention queries and keys rather than in the values or final output features; (2) output features semantically collapse, merging adjacent same-class objects that the queries and keys keep distinct; (3) this instance signal is bimodal in depth, strongest at the shallowest and deepest encoder stages; and (4) this signal is driven predominantly by the rotary position encoding (RoPE), whose removal collapses its advantage. We put these findings into our method TokenGraph3D, a training-free segmenter that groups points via connected components on a key-similarity graph, using neither density-based clustering nor proximity priors. Under identical prior-free conditions, we substantially outperform output-feature baselines, making the emergent 3D instance structure visible.
\end{abstract}

\vspace{-4.5mm}
\begin{figure*}[h]
    \centering
    \includegraphics[width=\textwidth]{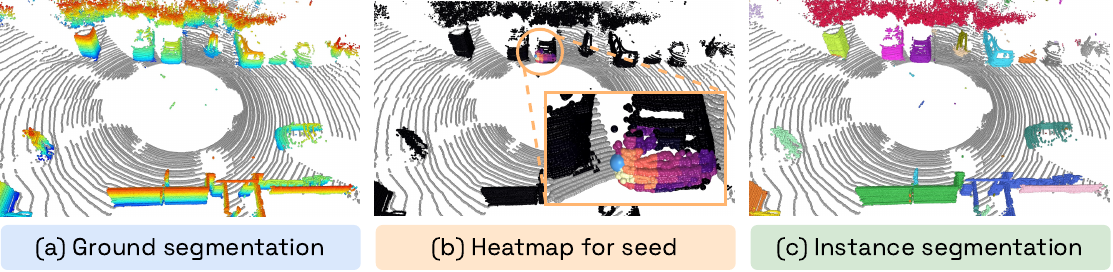}
    \caption{\textbf{Emergent 3D instance segmentation from a frozen, self-supervised point transformer.}
    (a) Outdoor LiDAR scan after ground removal via TerraSeg~\cite{lentsch2026terraseg}.
    (b) Cosine similarity of every point's attention key to the key of a single seed point (blue dot).
    (c) TokenGraph3D instance segmentation: connected components on the key-similarity graph, without density or proximity priors, recover the individual objects.
    }
    \label{fig:teaser}
\end{figure*}

\section{Introduction}\label{sec:intro}

Class-agnostic 3D instance segmentation, \ie partitioning a LiDAR scan into the individual objects it contains, without a fixed taxonomy, underpins autonomous driving, robotics, and large-scale auto-labeling. Learning it with full supervision requires dense, manually annotated point clouds that are costly to produce and tied to a closed set of classes. Unsupervised object discovery promises to sidestep this bottleneck by extracting object structure directly from raw sensor data.

However, essentially all existing unsupervised pipelines for outdoor LiDAR replace human labels with \emph{handcrafted geometric priors}, each resting on assumptions that do not hold in general. The dominant family removes the ground and then generates proposals by \emph{density-based spatial clustering}, such as DBSCAN~\cite{ester1996density} or HDBSCAN~\cite{mcinnes2017hdbscan}. Density clustering assumes a roughly homogeneous point density, but LiDAR density varies by \emph{several orders of magnitude} with range, \eg a nearby car can return thousands of points and a distant one only a handful, so a single neighborhood radius either fragments far objects or merges near ones. A second family relies on \emph{motion}, mining dynamic objects across repeated traversals or scene flow~\cite{you2022learning, baur2024liso}; this requires multiple scans and cannot discover static objects. A third projects \emph{2D detections} or features from synchronized cameras into 3D~\cite{lentsch2024union, khurana2025shelf}, inheriting calibration, synchronization, and field-of-view mismatches. In each case, the discovery signal comes from a geometric heuristic.

In 2D computer vision, a distinct paradigm has emerged. Self-supervised vision transformers (\eg DINO~\cite{caron2021emerging}) learn representations whose attention maps exhibit clean foreground-background separation, and works, such as LOST~\cite{simeoni2021localizing} and TokenCut~\cite{wang2022tokencut}, extract object masks directly from the network's \emph{attention keys} without retraining. Whether a frozen \emph{3D point transformer} exhibits the same emergent object structure, and what mechanism produces it, remains untouched by the literature. In this paper, we investigate three research questions:

\smallskip
\noindent\textbf{RQ1:} Do the internal attention representations of a self-supervised point transformer carry a stronger instance-level signal than its final output features?

\noindent\textbf{RQ2:} How does this instance-level signal vary with encoder depth?

\noindent\textbf{RQ3:} What mechanism (\eg positional encoding vs. raw token content) drives the emergence of this signal?
\smallskip

We take an \emph{analysis-first approach}. Treating the self-supervised point transformer Utonia~\cite{zhang2026utonia} as a frozen feature extractor, and \emph{starting from random seeds}, we measure how well each internal representation (\ie its queries, keys, and values across all encoder stages) isolates ground-truth objects, deliberately omitting any spatial proximity or clustering so that only representation quality is tested. Across the SemanticKITTI~\cite{behley2019semantickitti}, Panoptic nuScenes~\cite{fong2022panoptic, caesar2020nuscenes}, and Waymo Perception~\cite{sun2020scalability} datasets, four insights arise: (1) the instance signal concentrates in the attention queries and keys, not the values or output feature; (2) output features collapse on large objects, merging adjacent same-class neighbors that the keys keep distinct; (3) the signal is bimodal in depth, peaking at the shallowest and deepest stages; and (4) the signal is carried predominantly by the rotary position encoding (RoPE), without which the keys are no longer descriptive for separating instances. We operationalize these findings in \textbf{TokenGraph3D}, a training-free segmenter that groups points via connected components on a cosine-similarity graph over the emergent keys, with no density-based clustering and no proximity prior. See Fig.~\ref{fig:teaser}. In summary, our main contributions are:
\begin{enumerate}
    \item \textbf{Discovery of emergent 3D instance segmentation.} Using a geometric prior-free \emph{probing} protocol, we demonstrate that a self-supervised point transformer's internal queries and keys possess intrinsic instance awareness that outperforms its final output features. Mapping this signal across encoder depths and object sizes establishes a 3D counterpart to the 2D phenomenon.
    \item \textbf{Analysis of working mechanism.} We isolate three properties governing this emergent signal: (i) output features suffer from a \emph{scale-dependent semantic collapse}, merging adjacent same-class objects that keys keep distinct; (ii) the signal is bimodal in depth, strongest at the shallowest and deepest stages and weakest in between; and (iii) the signal is primarily \emph{positional rather than content-based}, driven by the rotary position encoding (RoPE).
    \item \textbf{TokenGraph3D.} We turn these findings into a training-free segmenter using only a self-supervised ground gate and connected components over the emergent key-similarity graph. Without density, motion, or 2D priors, it substantially outperforms the output-feature baseline and exceeds all published unsupervised methods on SemanticKITTI and nuScenes.
\end{enumerate}

\section{Related Work}\label{section:relatedwork}

We review object discovery in 2D using self-supervised vision transformers, survey the 3D LiDAR literature against which we position our work, and introduce the self-supervised models that enable our prior-free approach TokenGraph3D.

\subsection{Unsupervised 2D Object Discovery and Segmentation}
Self-supervised vision transformers (ViTs) underpin modern 2D unsupervised segmentation. Notably, the self-distillation objective of DINO~\cite{caron2021emerging} induces \emph{attention keys} that inherently capture clean object signals. Subsequent frameworks such as LOST~\cite{simeoni2021localizing} and TokenCut~\cite{wang2022tokencut} exploit this structure by constructing token-affinity graphs from these attention keys to segment objects without retraining or manual supervision. This line of work, extended to multi-object discovery via iterative spectral partitioning in CutLER~\cite{wang2023cut}, establishes a clear paradigm: leveraging frozen, self-supervised internal keys for graph-based grouping. We adapt this insight to 3D, \ie outdoor LiDAR, analyzing which representation holds the instance signal without relying on proximity or density priors.

\textbf{Nontrivial extension.}
Standard 2D approaches run global operations, \eg Normalized Cut~\cite{shi2000normalized}, over a centralized pool of a few hundred tokens~\cite{wang2022tokencut}. In contrast, Point Transformer V3 (PTv3)~\cite{wu2024ptv3}, the architecture of Utonia~\cite{zhang2026utonia}, utilizes \emph{patch attention}, where points are serialized along localized space-filling curves and grouped into non-overlapping spatial patches. Lacking a global \texttt{CLS} token or a single scene-wide attention matrix, PTv3 features are local, multi-pattern (varying serialization by depth), and multi-scale. Rather than attempting to reconstruct a massive global token graph, our TokenGraph3D extracts per-point key embeddings directly, constructing a simple point-level cosine affinity graph grouped via connected components. This architecture-compatible choice ensures downstream segmentation directly reflects representation quality.

\subsection{Unsupervised 3D Object Discovery and Segmentation}

Unsupervised LiDAR methods replace human annotations with handcrafted geometric heuristics, categorized by sensor modality and output representation.

\textbf{Unsupervised object discovery.} 
A prominent class of LiDAR-only methods relies on temporal motion cues, utilizing geometric ground removal, \eg RANSAC~\cite{fischler1981random}, followed by spatial clustering, such as DBSCAN~\cite{ester1996density} or HDBSCAN~\cite{mcinnes2017hdbscan}, to isolate moving objects across sequential frames~\cite{baur2024liso, najibi2022motion, you2022learning, zhang2023towards}. To discover static entities, multi-modal frameworks project self-supervised or weakly-supervised 2D features, \eg DINOv2~\cite{oquab2024dinov2}, into 3D space to separate foreground geometry~\cite{lentsch2024union, vobecky2022drive}. However, these methods have clear downsides: motion-based pipelines require expensive self-training to find static objects, which reduces precision, while multi-modal strategies suffer from sensor calibration errors or reintroduce human bias through pre-trained 2D models~\cite{kirillov2023segment, liu2024grounding}.

\textbf{Unsupervised instance segmentation.} 
Close to our objective are methods targeting class-agnostic, point-wise instance masks from LiDAR scans~\cite{nunes2022unsupervised, perauer2024autoinst, sautier2025unit}. However, these pipelines are tied to sparse convolutional backbones, such as MinkowskiNet~\cite{choy20194d}. This forces them to rely heavily on HDBSCAN, either to generate proposals on the fly (3DUIS~\cite{nunes2022unsupervised}) or to create pseudo-labels for training (AutoInst~\cite{perauer2024autoinst, nunes2023temporal} and UNIT~\cite{sautier2025unit}). Because these networks do not have internal attention mechanisms, they can only extract token similarities from the very final output features, leaving the power of their internal layers unused.
 
\textbf{Contributions.} 
TokenGraph3D differs in three main ways: (i) it pulls raw attention keys straight from a frozen, self-supervised point transformer instead of using encoder output embeddings; (ii) it works on single, isolated LiDAR scans without needing to aggregate multiple frames together; and (iii) it completely drops density-based clustering and proximity constraints when grouping points.

\subsection{Self-Supervised Point Transformers}

Two recent label-free models make our study possible, and we use both frozen. Utonia~\cite{zhang2026utonia} is a 137M-parameter point transformer~\cite{wu2024ptv3}, pretrained without labels on a large, heterogeneous mix of indoor and outdoor point clouds, built explicitly as a single scan encoder meant to transfer across sensors and domains; we extract its per-point, RoPE-augmented attention tokens for our analysis. TerraSeg~\cite{lentsch2026terraseg} provides a domain-agnostic, self-supervised ground gate. It has been trained with curated LiDAR data from 12 major public autonomous driving datasets. By combining these two foundation models, TokenGraph3D runs on any LiDAR scan completely \emph{free} of human labels and handcrafted spatial priors.

\section{Methodology}\label{sec:method}

\begin{figure*}[!b]
    \centering
    \includegraphics[width=\textwidth]{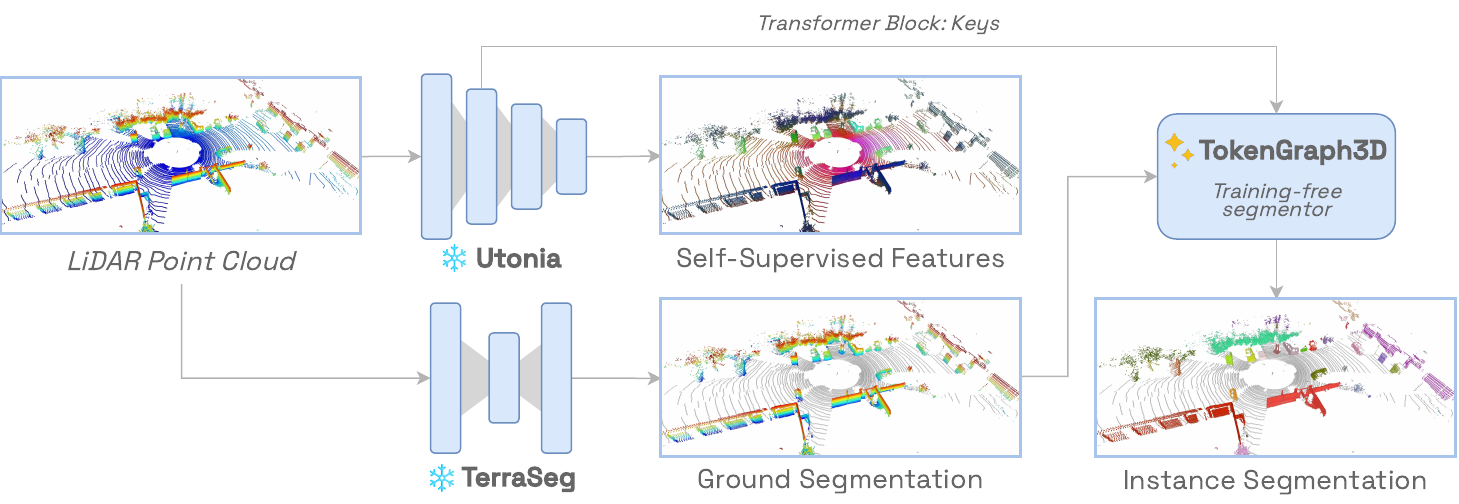}
    \caption{\textbf{TokenGraph3D.} A frozen, self-supervised point transformer (Utonia~\cite{zhang2026utonia}) provides per-point representations, \ie internal attention keys or output features, and TerraSeg~\cite{lentsch2026terraseg} removes the ground. Instances are the connected components of a cosine-affinity graph over the non-ground points, with no density and no proximity prior.}
    \label{fig:overview}
\end{figure*}

A point transformer trained by self-supervision optimizes a \emph{scan-level} objective, \ie masked reconstruction or contrastive alignment. To satisfy that objective, it must capture how points relate: which points share appearance, move together, or co-occur under augmentation. Hence, we hypothesize that these self-supervised learning objectives force the network to encode, implicitly and without any instance labels, an \emph{object-level} grouping: which points belong to the same physical entity (\eg a car or a pedestrian). If the hypothesis holds, that grouping should be readable from the frozen network's internal representations.

\textbf{TokenGraph3D.} Our method is a training-free instance segmenter assembled entirely from frozen, self-supervised models; it uses no instance labels, no 2D supervision, no motion cues, and, in its headline form, no density-based clustering and no geometric proximity prior. Every variant shares three stages (Fig.~\ref{fig:overview}): (i) a single forward pass of the frozen Utonia~\cite{zhang2026utonia} over the full LiDAR scan yields a per-point representation; (ii) the self-supervised ground segmenter TerraSeg~\cite{lentsch2026terraseg} removes ground points; and (iii) the remaining points are connected into a graph, with an edge linking two points when their features are sufficiently cosine-similar, and each connected component is reported as one instance. Variants differ only in \emph{which} per-point feature defines the edges, and we name them accordingly: by default, TokenGraph3D uses Utonia's internal post-RoPE \emph{attention keys} (Sec.~\ref{sec:method-keys}), while TokenGraph3D-Feat denotes the ablated variant that instead uses the \emph{output feature} embedding for connected components.

\subsection{Self-Supervised Per-Point Representations}\label{sec:method-keys}

All variants extract their per-point vectors from a \emph{single} forward pass of Utonia~\cite{zhang2026utonia}. This pass is executed over the \emph{entire} LiDAR scan, including ground points, to prevent an input distribution shift as the encoder was pretrained on complete scenes. Consequently, the ground is filtered only downstream (Sec.~\ref{sec:method-ground}). Regardless of which internal or output vector $\mathbf{g}_i$ a specific variant selects for point $i$, we apply $\ell_2$-normalization to obtain $\hat{\mathbf{g}}_i$. We use $\hat{\mathbf{g}}_i$ as the edge feature (Sec.~\ref{sec:method-graph}). The variants differ exclusively in their choice of the source vector $\mathbf{g}_i$.

\textbf{Probing.}
The hypothesis predicts that an \emph{object-level} grouping is already present in \emph{some} internal representation, but not necessarily in the output feature embedding that downstream heads normally consume. To locate it, we evaluate each candidate representation, \ie the queries, keys, and values at every stage together with the output feature, by how well it \emph{alone} separates a ground-truth object from the rest of the scan. Crucially, we withhold every grouping mechanism: no spatial proximity and no density-based clustering. A representation therefore scores well only if its own embedding content, and nothing else, isolates the object, which lets us compare representations and pinpoint where the emergent grouping lives before we commit the segmenter to any one of them. We quantify this with a seed-based overlap score, defined in Sec.~\ref{sec:metrics}.

\textbf{Attention keys.}
Guided by the probe, TokenGraph3D is constructed using post-RoPE attention keys extracted from a single forward pass of a chosen encoder block. Because deep stages pool 3D points into sparse grid tokens, every non-ground point inherits the key of its corresponding token, \ie multiple points can have the same key embedding. This formulation ensures that points within the same token remain grouped, while points across different tokens are compared via key cosine similarity. Two architectural design choices define this variant. First, for the encoder stage, we default to the deep \texttt{enc4} block. Segmentation quality is bimodal in depth (Sec.~\ref{sec:results}): the shallowest (\texttt{enc0}) and deepest (\texttt{enc4}) keys are the strongest and roughly tied, while the intermediate stages are weaker. We adopt \texttt{enc4} because it reaches this quality with far fewer, cleaner clusters at a robust threshold, whereas the shallow \texttt{enc0} key needs a near-maximal threshold and heavily over-segments; \texttt{enc4}'s coarse tokens also resist the transitive over-merging that connected components suffer without a proximity gate. Second, regarding the choice of keys over queries, while both perform comparably within the probe, keys yield marginally stronger performance at our operating stage and align with established 2D object-discovery conventions. We use post-RoPE keys by default and evaluate \emph{pre-RoPE} keys only as an ablation. Our findings show that the instance signal is driven by positional rotation, not raw key content. For comparison, the ablated TokenGraph3D-Feat bypasses the attention blocks and takes the backbone's final output feature directly, $\mathbf{g}_i = \mathbf{f}_i \in \mathbb{R}^{C}$.

\subsection{Self-Supervised Ground Gate}\label{sec:method-ground}

We utilize TerraSeg~\cite{lentsch2026terraseg} to partition each scan into ground and non-ground points. Only non-ground points are grouped into instances, while ground points receive a ``no prediction'' label. Crucially, this filtering step is a purely \emph{semantic} classification (\ie ground versus non-ground class) and does not dictate how non-ground points are clustered into distinct objects; it therefore introduces no geometric instance prior. Ground removal is standard in unsupervised LiDAR pipelines for both object detection and instance discovery, \eg LISO~\cite{baur2024liso} and UNION~\cite{lentsch2024union}, where it is typically realized using heuristic planar fits such as RANSAC~\cite{fischler1981random}. Because TerraSeg is itself self-supervised, it replaces these rigid geometric heuristics with a learned gate, keeping our pipeline human-label-free.

\subsection{Point Grouping}\label{sec:method-graph}

\textbf{Feature-only grouping.}
Over the non-ground points, we build an undirected graph. An edge connects points $i$ and $j$ if and only if their edge features are sufficiently aligned, \ie their cosine similarity exceeds a fixed threshold $\tau \in [0,1)$:
\begin{equation}
\langle \hat{\mathbf{g}}_i, \hat{\mathbf{g}}_j \rangle > \tau,
\label{eq:edge-free}
\end{equation}
where $\langle \hat{\mathbf{g}}_i, \hat{\mathbf{g}}_j \rangle \in [-1,1]$ is the cosine similarity of the $\ell_2$-normalized edge features, \ie the attention key for TokenGraph3D. Each instance is a connected component of this graph: two points belong to the same object if a path of feature-coherent edges joins them. No spatial distance or point density enters the decision, so grouping is driven entirely by the representation learned through self-supervision. In this regime, nothing but feature quality separates objects, so it is precisely where the hypothesized emergent structure must reveal itself.

\textbf{Proximity-gated grouping [ablation].}
To test the value of a geometric prior, a hard proximity gate admits an edge only when both conditions hold:
\begin{equation}
    \lVert \mathbf{p}_i - \mathbf{p}_j \rVert_2 \le r_{\max}
    \quad\text{and}\quad
    \langle \hat{\mathbf{g}}_i, \hat{\mathbf{g}}_j \rangle > \tau .
    \label{eq:edge}
\end{equation}
The first term is a proximity constraint of radius $r_{\max}$; the second is the same feature criterion as in Eq.~\eqref{eq:edge-free}. This cleanly separates the two roles distance plays in density clustering: geometry only \emph{gates} which points $p_i$ may be compared, while the decision to connect them rests entirely with the \emph{learned features}. This variant solely quantifies the prior's cost (ablation); it is not our proposed method.

\textbf{No density threshold.} In both variants, every connected component is reported as an instance: there is no \texttt{min\_cluster\_size} hyperparameter and no density-based filtering. This keeps the method from collapsing into ``DBSCAN with learned features'', and it avoids a tuned density which, because LiDAR point density varies strongly with range, tends not to transfer across sensors with different beam count, \eg sparse vs.\ dense LiDAR scans.

\subsection{Hyperparameters}\label{sec:method-hparams}
The feature-only method has a single hyperparameter, the cosine threshold $\tau$; the proximity-gated ablation adds a radius $r_{\max}$. The remaining choices, the encoder stage and whether to read keys, queries, or the output feature, are settled by our analysis (Sec.~\ref{sec:results}) rather than hand-tuned. The method inherits its cross-sensor robustness from the two frozen, multi-domain self-supervised models, which is why we expect a single configuration to transfer across LiDAR of differing beam density. The pipeline treats these models as interchangeable modules that can be substituted whenever stronger alternatives become available in the future.

\section{Experiments}\label{sec:experiments}

We first describe our implementation details, datasets, metrics, and baselines. After that, we report three results: a representation probe that localizes the emergent instance signal, the prior-free segmentation result in which that signal becomes visible, and a cost-of-prior comparison against unsupervised baselines.

\subsection{Implementation Details}\label{sec:implementation}

Both point transformers are kept completely frozen without fine-tuning or downstream adaptation: Utonia ($137\text{M}$ parameters) extracts per-point features at coordinate scale $0.5$, while TerraSeg (variant B, $46\text{M}$ parameters) serves as the ground gate. Each LiDAR point cloud is transformed into TerraSeg's canonical ego frame (ground near $z=0$, $+x$ forward). We process each scan independently, without temporal aggregation or semantic supervision at inference. Our method relies on a single cosine threshold $\tau$ (and, for proximity ablations, a maximum radius $r_{\max}$), tuned once on pooled training data and held fixed across validation sets (Sec.~\ref{sec:baselines}). We release our code on \href{https://github.com/TedLentsch/TokenGraph3D}{GitHub} under the Apache 2.0 license.

\subsection{Datasets}\label{sec:datasets}

We evaluate on three outdoor LiDAR benchmarks that differ in sensor resolution and scene geometry. This diversity lets us test whether a single configuration transfers across diverse conditions rather than overfitting one sensor.

\textbf{SemanticKITTI~\cite{behley2019semantickitti}.}
It has dense $64$-beam Velodyne HDL-64E scans at \qty{10}{\hertz} in urban and highway scenes. We follow the official split, tuning on the training sequences ($00$--$07$, $09$, $10$) and evaluating on the held-out validation sequence ($08$); point-wise instance IDs are provided for the \emph{thing} classes.

\textbf{Panoptic nuScenes~\cite{caesar2020nuscenes,fong2022panoptic}.}
A multi-modal driving dataset of $1000$ \qty{20}{\second} scenes from Boston and Singapore, captured with a sparse $32$-beam LiDAR and annotated with point-wise panoptic labels on the keyframes at \qty{2}{\hertz}. We use the official \texttt{train}/\texttt{val} split and operate on the annotated keyframes. Its lower resolution and different scene geometry make it complementary to KITTI.

\textbf{Waymo Perception~\cite{sun2020scalability}.}
A large-scale driving dataset recorded with a $64$-beam LiDAR across diverse cities and conditions. The dataset has $798$ train sequences, $202$ validation sequences, and $150$ test sequences. It provides 3D bounding boxes but no point-wise instance labels, so we derive panoptic segmentation instances from the boxes: each foreground point takes the instance of the box that contains it, and foreground points inside no box are left unscored, as their instance is undefined. These box-derived labels are coarser than manual panoptic labels, making Waymo a challenging test of cross-sensor transfer.

\subsection{Metrics}\label{sec:metrics}

We evaluate class-agnostic instance segmentation using the association score $S_{\mathrm{assoc}}$~\cite{nunes2022unsupervised, sautier2025unit} in the single-scan regime. For a ground-truth instance $t$ and a predicted segment $s$, the intersection-over-union ($\mathrm{IoU}$) is
\begin{equation}
    \mathrm{IoU}(s,t) = \frac{|s\cap t|}{|s| + |t| - |s\cap t|},
    \label{eq:iou}
\end{equation}
and $S_{\mathrm{assoc}}$ aggregates this over all predictions overlapping each target instance:
\begin{equation}
    S_{\mathrm{assoc}} = \frac{1}{|\mathcal{T}|} \sum_{t\in\mathcal{T}} \frac{1}{|t|} \sum_{s:\,|s\cap t|>0} |s\cap t|\;\mathrm{IoU}(s,t),
    \label{eq:sassoc}
\end{equation}
where $\mathcal{T}$ denotes the set of scored ground-truth instances. The metric rewards predictions that tightly encapsulate an object without leaking into surrounding geometry, while the normalization by $|t|$ prevents large objects from dominating the score. Following~\cite{sautier2025unit}, we score each scan independently and average across scans. Stuff points are explicitly retained during evaluation; predictions that bleed onto background classes are penalized via an inflated denominator $|s|$.

\textbf{Representation probe (seed-IoU).}
To quantify the grouping quality of raw representations independently of any clustering algorithm or graph heuristics, we introduce a scaffolding-free probe. For each ground-truth instance, we uniformly sample up to $50$ seed points across the object. From each seed, we select all scan points whose cosine similarity exceeds a threshold and measure the $\mathrm{IoU}$ between this selected set and the ground-truth object. We average over all seeds and instances, evaluating each representation at its optimal threshold. Because this metric omits spatial radius constraints and connected-component filtering, it isolates how cleanly the feature space alone separates an object from the rest of the scan. Seed-IoU is strictly diagnostic and serves to rank internal representations within this probe.

\subsection{Baselines and Tuning Protocol}\label{sec:baselines}

We benchmark against two categories of unsupervised methods:

\textbf{Unsupervised pipelines.}
We compare against the unsupervised LiDAR instance segmentation literature: 3DUIS~\cite{nunes2022unsupervised}, TARL-Seg~\cite{nunes2023temporal}, 4D-Seg~\cite{sautier2025unit}, and UNIT~\cite{sautier2025unit}. Following~\cite{sautier2025unit}, we report their published results directly.

\textbf{Spatial clustering baselines.}
To isolate the contribution of emergent feature representations from pure spatial grouping, we evaluate DBSCAN~\cite{ester1996density} and HDBSCAN~\cite{mcinnes2017hdbscan} using their efficient implementations in TorchRobotics~\cite{lentsch2025torchrobotics}. To ensure a strictly controlled comparison, both share our identical TerraSeg~\cite{lentsch2026terraseg} ground-gating module and operate on the exact same non-ground points, differing from our approach solely in the grouping mechanism.

\textbf{Cross-dataset tuning protocol.}
To test genuine generalization and prevent dataset overfitting, every method is tuned to obtain the single best configuration across all three benchmarks (Sec.~\ref{sec:datasets}). We perform a grid search on a pooled $250$-scan training subset to maximize mean $S_{\mathrm{assoc}}$, then evaluate this fixed configuration on the validation splits.
Specifically, we search cosine threshold $\tau \in [0.90, 0.99]$ for TokenGraph3D (alongside radius $r_{\max} \in [0.5, 1.25]$\,m for proximity ablations), radius $\varepsilon \in [0.5, 1.5]$\,m and \texttt{min\_samples} $\in [1, 25]$ for DBSCAN, and \texttt{min\_cluster\_size} $\in [2, 25]$ for HDBSCAN.

\begin{figure}[t]
    \centering
    \includegraphics[width=\linewidth]{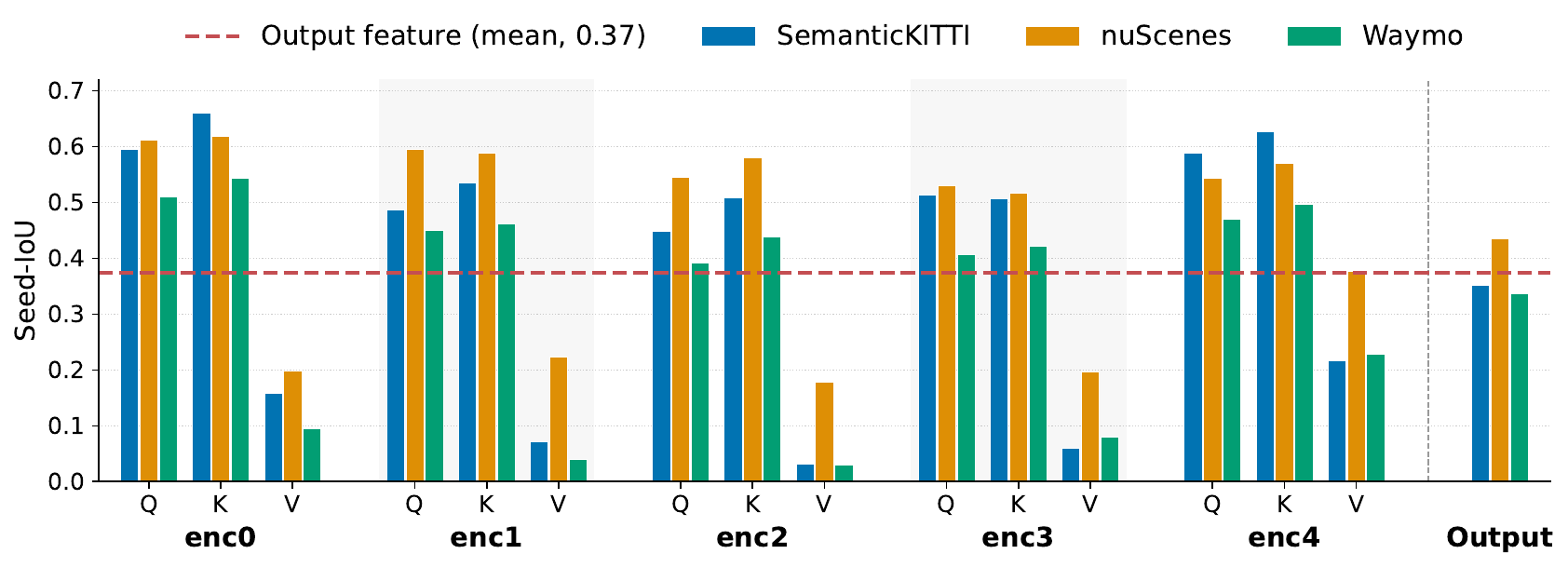}
    \caption{
    \textbf{Representation probe [Finding 1].}
    For each encoder stage, we compare the query, key, and value against the output feature, on all three datasets. Across every stage and dataset the query and key carry a strong, nearly identical instance signal, the value is weak, and the output feature falls below the attention queries and keys.
    }
    \label{fig:probe}
\end{figure}

\subsection{Probing: Localizing Instance Signal}\label{sec:results}

We run the \emph{probe} over the queries, keys, and values at each encoder stage (all heads concatenated), together with the output feature, on the training split of each dataset. Pooling over all foreground instances makes the rankings below stable. Four findings hold across every dataset.

\vspace{1mm}
\finding{Finding 1: The instance signal concentrates in the queries and keys, not in the values or the output feature.}

\noindent At every encoder stage and on every dataset, the queries and keys score above the average output-feature seed-IoU (Fig.~\ref{fig:probe}). The margin is widest at the two ends of the network: at \texttt{enc0} and \texttt{enc4}. The query and key seed-IoU, averaged over the three datasets, exceeds the averaged output feature score by more than $0.10$. Queries track keys almost exactly throughout, while values are the weakest projection, below even the output feature. The object signal therefore lives in the query-key matching space rather than in the aggregated value or output representation. This is the 3D counterpart of the 2D key-space discovery phenomenon~\cite{simeoni2021localizing, wang2022tokencut}, with the twist that here queries are as informative as keys.

\vspace{1mm}
\finding{Finding 2: The output feature collapses on large objects, while the key stays robust.}

\begin{figure}[t]
    \centering
    \includegraphics[width=\linewidth]{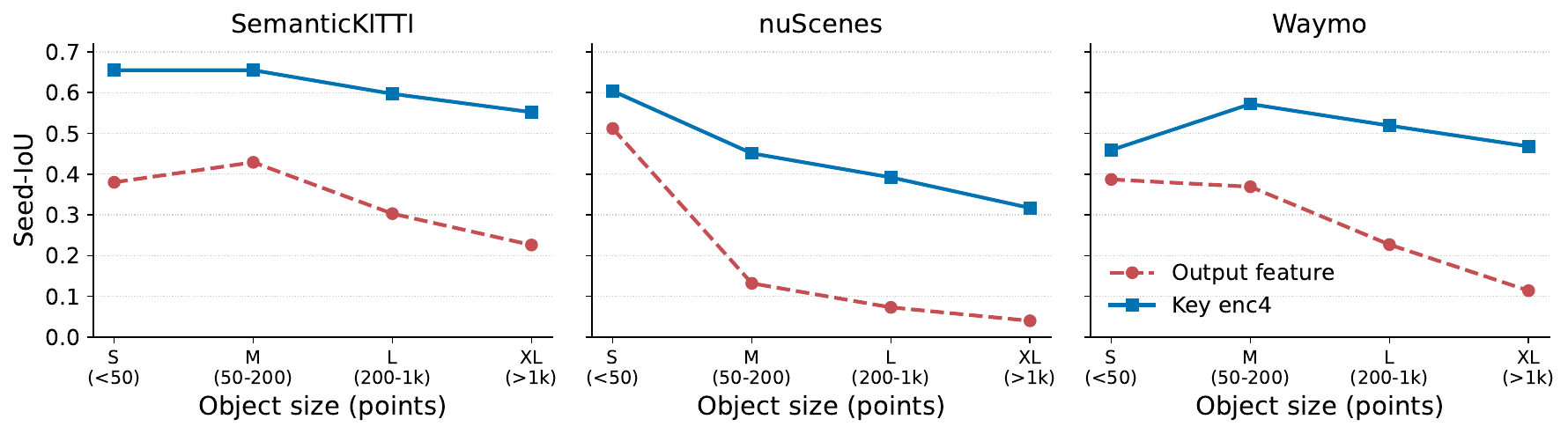}
    \caption{
    \textbf{The output feature's seed-IoU collapses on large objects [Finding~2].}
    Mean seed-IoU versus object size (number of non-ground points), for the output feature and the \texttt{enc4} attention key, each at its own best cosine threshold. As objects grow the output feature loses performance, whether by merging neighboring objects or capturing only a fragment, while the attention key remains discriminative on all three datasets. The effect is strongest for the relatively sparse LiDAR of the nuScenes~\cite{caesar2020nuscenes} dataset.
    }
    \label{fig:size}
\end{figure}

\noindent Grouping foreground objects by their point count (Fig.~\ref{fig:size}), the output feature seed-IoU decreases from small to large objects on all three datasets, for example from $0.38$ to $0.23$ on SemanticKITTI, whereas the key remains high and nearly flat. The output feature thus loses large objects, consistent with either merging same-class neighbors or covering only a fragment; its strong under-segmentation (Tab.~\ref{tab:proxfree}: only $193.9$ vs.\ $465.9$ clusters per scan for the key on nuScenes) points to merging. We use point count as a proxy for object size; because LiDAR density falls with range, this proxy conflates physical size with density, a caveat we return to in the limitations part of the Conclusion (Sec.~\ref{section:conclusion}).

\vspace{1mm}
\finding{Finding 3: The instance signal is bimodal in depth, strongest at the shallowest and deepest stages.}

\noindent Prior-free segmentation quality is U-shaped across encoder stages (Tab.~\ref{tab:proxfree}): the shallowest (\texttt{enc0}) and deepest (\texttt{enc4}) keys are the strongest and essentially tied (average $S_{\mathrm{assoc}}$ $0.632$ vs.\ $0.630$), while the intermediate stages are markedly weaker (\texttt{enc3} is the lowest at $0.524$). The same U-shape appears in the seed-IoU probe (Fig.~\ref{fig:probe}). We default to \texttt{enc4} rather than \texttt{enc0}: it reaches the same quality with far fewer clusters (nuScenes: $466$ vs.\ $1364$ per scan) at a more robust threshold ($\tau=0.95$), whereas \texttt{enc0} peaks only at a near-maximal $\tau=0.99$ and over-segments, and \texttt{enc4}'s coarser tokens further resist the transitive over-merging that connected components suffer without a radius (\ie proximity prior).

\vspace{1mm}
\finding{Finding 4: The rotary position encoding carries the instance signal.}

\noindent Removing the rotary position encoding (RoPE) collapses the key signal at every stage. In the probe, pre-RoPE keys fall to $0.01$--$0.05$ seed-IoU in the shallow stages, and at the deepest stage a pre-RoPE key is no more discriminative than a value. The collapse carries over to segmentation, where the pre-RoPE \texttt{enc4} key falls below even the output feature (Tab.~\ref{tab:proxfree}, Pre-RoPE row). RoPE, not the raw key content, is therefore what lifts keys above values, and we use post-RoPE keys in this work.
Since RoPE injects positional structure, one might read this as geometry re-entering through the representation; prior-free in this paper, however, means no \emph{explicit handcrafted} geometric rule (no radius, no density, no motion or 2D cue). That the frozen network turns raw position into a discriminative instance signal is not a confound but precisely the emergent capability under study.

\subsection{Connected Components: Prior-Free Instance Segmentation}\label{sec:proxfree-results}

We first evaluate \emph{TokenGraph3D} in its headline, prior-free form: connected components on a feature-only similarity graph, with no proximity and no density prior (Eq.~\eqref{eq:edge-free}), varying only the per-point feature (Tab.~\ref{tab:proxfree}). Under these identical conditions, the emergent keys segment far better than the output feature, turning the representation gap of Sec.~\ref{sec:results} into a segmentation gap. The \texttt{enc4} key reaches $0.726$, $0.517$, and $0.646$ $S_{\mathrm{assoc}}$ on SemanticKITTI, nuScenes, and Waymo, against $0.486$, $0.273$, and $0.455$ for the output feature, a gain of $+0.24$, $+0.24$, and $+0.19$. The only stage that approaches \texttt{enc4} is the shallow \texttt{enc0} key (Finding 3), and queries carry nearly the same signal as keys (Fig.~\ref{fig:probe}). The nuScenes output feature is the clearest case: it collapses to $0.273$ while the \texttt{enc4} key holds at $0.517$, exactly where prior-free grouping is hardest, on small, sparse objects, and it does so by merging them, as reflected in its lower cluster count ($193.9$ vs.\ $465.9$ per scan; Tab.~\ref{tab:proxfree}). This supports our claim: the instance structure is emergent in the keys and becomes visible once there is no geometric prior. Even prior-free, TokenGraph3D already outperforms every method on SemanticKITTI ($0.726$ vs.\ UNIT $0.715$), surpasses all published unsupervised methods on nuScenes ($0.517$ vs.\ $0.390$), and matches density-based clustering on Waymo (Tables~\ref{tab:semantickitti}--\ref{tab:waymo}).

\begin{table}[t]
    \centering
    \small
    \setlength{\tabcolsep}{4pt}
    \caption{
    \textbf{Prior-free instance segmentation.}
    Association score $S_{\mathrm{assoc}}$ (higher is better) and average predicted clusters per scan (\#clust.) on each validation split, under the shared prior-free pipeline (Eq.~\eqref{eq:edge-free}). We list the two peak stages, \texttt{enc0} and \texttt{enc4}; the intermediate stages are weaker (Finding~3). Best in \textbf{bold}, second-best \underline{underlined}.
    }
    \label{tab:proxfree}
    \begin{tabular}{l c cc cc cc}
        \toprule
        & & \multicolumn{2}{c}{SemanticKITTI} & \multicolumn{2}{c}{nuScenes} & \multicolumn{2}{c}{Waymo Perception} \\
        \cmidrule(lr){3-4}\cmidrule(lr){5-6}\cmidrule(lr){7-8}
        Node feature & $\tau$ & $S_{\mathrm{assoc}}$ & \#clust. & $S_{\mathrm{assoc}}$ & \#clust. & $S_{\mathrm{assoc}}$ & \#clust. \\
        \midrule
        \texttt{enc0} Key & 0.99 & \underline{0.713} & 723.6 & \textbf{0.534} & 1363.5 & \textbf{0.650} & 532.9 \\
        \texttt{enc3} Key & 0.95 & 0.571 & 864.1 & 0.467 & 539.2 & 0.533 & 1022.4 \\
        \texttt{enc4} Key & 0.95 & \textbf{0.726} & 434.6 & \underline{0.517} & 465.9 & \underline{0.646} & 545.0 \\
        \;\;\textit{Pre-RoPE} & 0.98 & 0.236 & 1487.3 & 0.141 & 255.3 & 0.191 & 2143.3 \\
        \texttt{enc4} Query & 0.97 & 0.680 & 479.2 & 0.485 & 377.3 & 0.627 & 573.6 \\
        \midrule
        Output Embedding & 0.95 & 0.486 & 518.8 & 0.273 & 193.9 & 0.455 & 428.3 \\
        \bottomrule
    \end{tabular}
\end{table}

\begin{table}[t]
    \centering
    \small
    \setlength{\tabcolsep}{2.5pt}
    \caption{
    \textbf{Geometric prior on SemanticKITTI validation split.}
    We report the association score $S_{\text{assoc}}$ and the average number of predicted clusters per scan, using the same pipeline as Tab.~\ref{tab:proxfree}. $^\dagger$Taken from~\cite{sautier2025unit}. Best in \textbf{bold}, second-best is \underline{underlined}.
    }
    \label{tab:semantickitti}
    \begin{tabular}{l ccc cc}
        \toprule
        \multirow{2}{*}{Method} & \multicolumn{3}{c}{Heuristics and Priors} & \multicolumn{2}{c}{Metrics} \\
        \cmidrule(l{2pt}r{2pt}){2-4} \cmidrule(l{2pt}r{2pt}){5-6}
        & No 2D & No temporal & No density & $S_{\text{assoc}}$ & \#clusters \\
        \midrule
        TerraSeg~\cite{lentsch2026terraseg} + DBSCAN~\cite{ester1996density} & \textcolor{citationblue}{\cmark} & \textcolor{citationblue}{\cmark} & \textcolor{gray}{\xmark} & 0.708 & 500.1 \\
        TerraSeg~\cite{lentsch2026terraseg} + HDBSCAN~\cite{mcinnes2017hdbscan} & \textcolor{citationblue}{\cmark} & \textcolor{citationblue}{\cmark} & \textcolor{gray}{\xmark} & 0.689 & 436.6 \\
        3DUIS~\cite{nunes2022unsupervised} & \textcolor{citationblue}{\cmark} & \textcolor{citationblue}{\cmark} & \textcolor{gray}{\xmark} & 0.550$^{\dagger}$ & - \\
        TARL-Seg~\cite{nunes2023temporal} & \textcolor{citationblue}{\cmark} & \textcolor{gray}{\xmark} & \textcolor{gray}{\xmark} & 0.668$^{\dagger}$ & - \\
        4D-Seg~\cite{sautier2025unit} & \textcolor{citationblue}{\cmark} & \textcolor{gray}{\xmark} & \textcolor{gray}{\xmark} & 0.667$^{\dagger}$ & - \\
        UNIT~\cite{sautier2025unit} & \textcolor{citationblue}{\cmark} & \textcolor{gray}{\xmark} & \textcolor{gray}{\xmark} & 0.715$^{\dagger}$ & - \\
        \midrule
        TokenGraph3D (ours) & \textcolor{citationblue}{\cmark} & \textcolor{citationblue}{\cmark} & \textcolor{citationblue}{\cmark} & 0.726 & 434.6 \\
        \;\;+ proximity (Key) & \textcolor{citationblue}{\cmark} & \textcolor{citationblue}{\cmark} & \textcolor{gray}{\xmark} & \textbf{0.747} & 563.3 \\
        \;\;+ proximity (Feat) & \textcolor{citationblue}{\cmark} & \textcolor{citationblue}{\cmark} & \textcolor{gray}{\xmark} & \underline{0.741} & 543.2 \\
        \bottomrule
    \end{tabular}
\end{table}

\begin{table}[!t]
    \centering
    \small
    \setlength{\tabcolsep}{2.5pt}
    \caption{
    \textbf{Geometric prior on nuScenes validation split.}
    We report the association score $S_{\text{assoc}}$ and the average number of predicted clusters per scan, using the same pipeline as Tab.~\ref{tab:proxfree}. $^\dagger$Taken from~\cite{sautier2025unit}. Best in \textbf{bold}, second-best is \underline{underlined}.
    }
    \label{tab:nuscenes}
    \begin{tabular}{l ccc cc}
        \toprule
        \multirow{2}{*}{Method} & \multicolumn{3}{c}{Heuristics and Priors} & \multicolumn{2}{c}{Metrics} \\
        \cmidrule(l{2pt}r{2pt}){2-4} \cmidrule(l{2pt}r{2pt}){5-6}
        & No 2D & No temporal & No density & $S_{\text{assoc}}$ & \#clusters \\
        \midrule
        TerraSeg~\cite{lentsch2026terraseg} + DBSCAN~\cite{ester1996density} & \textcolor{citationblue}{\cmark} & \textcolor{citationblue}{\cmark} & \textcolor{gray}{\xmark} & 0.544 & 1008.5 \\
        TerraSeg~\cite{lentsch2026terraseg} + HDBSCAN~\cite{mcinnes2017hdbscan} & \textcolor{citationblue}{\cmark} & \textcolor{citationblue}{\cmark} & \textcolor{gray}{\xmark} & 0.374 & 244.0 \\
        TARL-Seg~\cite{nunes2023temporal} & \textcolor{citationblue}{\cmark} & \textcolor{gray}{\xmark} & \textcolor{gray}{\xmark} & 0.189$^{\dagger}$ & - \\
        4D-Seg~\cite{sautier2025unit} & \textcolor{citationblue}{\cmark} & \textcolor{gray}{\xmark} & \textcolor{gray}{\xmark} & 0.287$^{\dagger}$ & - \\
        UNIT~\cite{sautier2025unit} & \textcolor{citationblue}{\cmark} & \textcolor{gray}{\xmark} & \textcolor{gray}{\xmark} & 0.390$^{\dagger}$ & - \\
        \midrule
        TokenGraph3D (ours) & \textcolor{citationblue}{\cmark} & \textcolor{citationblue}{\cmark} & \textcolor{citationblue}{\cmark} & 0.517 & 465.9 \\
        \;\;+ proximity (Key) & \textcolor{citationblue}{\cmark} & \textcolor{citationblue}{\cmark} & \textcolor{gray}{\xmark} & \underline{0.546} & 1037.2 \\
        \;\;+ proximity (Feat) & \textcolor{citationblue}{\cmark} & \textcolor{citationblue}{\cmark} & \textcolor{gray}{\xmark} & \textbf{0.547} & 1025.6 \\
        \bottomrule
    \end{tabular}
\end{table}

\begin{table}[t]
    \centering
    \small
    \setlength{\tabcolsep}{2.5pt}
    \caption{
    \textbf{Geometric prior on Waymo Perception validation split.}
    We report the association score $S_{\text{assoc}}$ and the average number of predicted clusters per scan, using the same pipeline as Tab.~\ref{tab:proxfree}. Best in \textbf{bold}, second-best is \underline{underlined}.
    }
    \label{tab:waymo}
    \begin{tabular}{l ccc cc}
        \toprule
        \multirow{2}{*}{Method} & \multicolumn{3}{c}{Heuristics and Priors} & \multicolumn{2}{c}{Metrics} \\
        \cmidrule(l{2pt}r{2pt}){2-4} \cmidrule(l{2pt}r{2pt}){5-6}
        & No 2D & No temporal & No density & $S_{\text{assoc}}$ & \#clusters \\
        \midrule
        TerraSeg~\cite{lentsch2026terraseg} + DBSCAN~\cite{ester1996density} & \textcolor{citationblue}{\cmark} & \textcolor{citationblue}{\cmark} & \textcolor{gray}{\xmark} & 0.636 & 401.9 \\
        TerraSeg~\cite{lentsch2026terraseg} + HDBSCAN~\cite{mcinnes2017hdbscan} & \textcolor{citationblue}{\cmark} & \textcolor{citationblue}{\cmark} & \textcolor{gray}{\xmark} & 0.648 & 537.4 \\
        \midrule
        TokenGraph3D (ours) & \textcolor{citationblue}{\cmark} & \textcolor{citationblue}{\cmark} & \textcolor{citationblue}{\cmark} & 0.646 & 545.0 \\
        \;\;+ proximity (Key) & \textcolor{citationblue}{\cmark} & \textcolor{citationblue}{\cmark} & \textcolor{gray}{\xmark} & \textbf{0.668} & 491.4 \\
        \;\;+ proximity (Feat) & \textcolor{citationblue}{\cmark} & \textcolor{citationblue}{\cmark} & \textcolor{gray}{\xmark} & \underline{0.662} & 431.3 \\
        \bottomrule
    \end{tabular}
\end{table}

\subsection{Ablation: Adding Geometric Prior}\label{sec:costofprior}

To separate how much of the segmentation quality comes from the representation and how much a geometric prior can supply, we now add back the weakest prior: a proximity gate that only lets nearby points be linked (Eq.~\eqref{eq:edge}). Tables~\ref{tab:semantickitti}--\ref{tab:waymo} compare this proximity-gated variant with the unsupervised baselines. The point of the comparison is that adding proximity makes the representation almost irrelevant: the Feat and key variants come within a few thousandths of each other and of DBSCAN (nuScenes: $0.547$ vs.\ $0.546$), so the prior-free representation gap of $0.24$ collapses to under $0.01$. The proximity constraint suppresses precisely the long-range wrong merges on which the representations differ; once geometry rules those out, correct merging no longer depends on the representation, and the score barely moves. Proximity also raises the predicted cluster count, since it forbids distant same-feature points from linking, although the per-method tuning of $\tau$ partly obscures this. This absorption of the representation advantage by geometry is exactly why the prior-free setting (Sec.~\ref{sec:proxfree-results}) is where the emergent signal must be read.

\section{Conclusion}\label{section:conclusion}

This paper set out to test a single hypothesis: that a frozen, self-supervised point transformer already encodes which points belong to the same object, with no handcrafted geometric prior. By probing its internal representations (the queries, keys, and values of each encoder stage, together with the output feature) using a proximity- and clustering-free metric, we confirm that it does. The attention keys and queries isolate instances far better than the output feature (Finding~1); the output feature collapses on large and adjacent objects while the keys hold (Finding~2); the signal is bimodal in depth, strongest at the shallowest and deepest stages (Finding~3); and it is created by the rotary position encoding and vanishes when RoPE is removed (Finding~4). Without a geometric prior the representation is decisive: on nuScenes the output feature reaches only $0.273$ $S_{\mathrm{assoc}}$ and predicts $193.9$ clusters per scan, merging distinct objects, whereas the \texttt{enc4} key nearly doubles it to $0.517$. Frozen keys thus already encode object structure, with no training, proximity, or density. TokenGraph3D operationalizes this as a training-free segmenter that groups points by connected components on the emergent key graph, with no density clustering, motion, or $2$D supervision.

\textbf{Limitations.}
Our aim is analysis rather than benchmarking. Although our prior-free TokenGraph3D already matches or exceeds existing unsupervised methods on single-scan $S_{\mathrm{assoc}}$, adding a proximity prior absorbs most of the representation advantage, so learned features then sit in the same territory as density clustering, such as DBSCAN~\cite{ester1996density}. Operating without proximity also has two predictable failure modes: the method can over-segment very large objects, visible in the high cluster counts of the shallow-key variants (Tab.~\ref{tab:proxfree}), and can merge distant objects of the same class through chains of feature-coherent edges. The proximity ablation (Sec.~\ref{sec:costofprior}) improves exactly these cases, which supports this interpretation. Moreover, all findings are established on a single self-supervised backbone, Utonia~\cite{zhang2026utonia}, and although the segmenter is training-free, the cosine threshold $\tau$ and the representation choice are selected on a labeled 250-scan subset (Sec.~\ref{sec:baselines}); probing supervised or alternative architectures, and a label-free selection rule, remain open. Finally, our scale analysis (Finding~2) uses point count as a proxy for object size; because LiDAR density decreases with range, this proxy partly conflates size with sparsity, and disentangling the two would require object-area annotations we do not use.

\textbf{Future work.}
The bimodal depth structure suggests fusing the shallow \texttt{enc0} and deep \texttt{enc4} keys, combining the fine localization of the former with the clean, robust grouping of the latter. A further direction is to test whether the same emergent key structure arises in other self-supervised point transformer models, such as Sonata~\cite{wu2025sonata}, and in volumetric transformers, such as Volt~\cite{yilmaz2026volt}.

\section*{Acknowledgements}
This work has been supported by project PID2024-161576OB-I00, funded by Spanish MICIU/AEI/10.13039/501100011033 and co-funded by the European Regional Development Fund (ERDF, ``A way of making Europe'').

\bibliographystyle{splncs04}
\bibliography{main}
\end{document}